\documentclass[final,5p,times]{elsarticle}

\usepackage{multirow}
\usepackage{amsmath}
\usepackage{graphicx}
\usepackage{natbib}
\usepackage{marvosym}
\usepackage{amssymb}
\usepackage{booktabs} 
\usepackage{multirow}
\usepackage{color}
\usepackage{hyperref}

\usepackage[linesnumbered,ruled]{algorithm2e}
\SetKwProg{Fn}{Function}{}{end}\SetKwFunction{FRecurs}{FnRecursive}%

\usepackage{etoolbox}

\usepackage{lineno}

\begin{document}

\begin{frontmatter}



\title{Embedding Push and Pull Search in the Framework of Differential Evolution for Solving Constrained Single-objective Optimization Problems}







\author[addr1]{Zhun Fan}
\author[addr1]{Wenji Li}
\author[addr1]{Zhaojun Wang}
\author[addr1]{Yutong Yuan}
\author[addr1]{Fuzan Sun}
\author[addr1]{Zhi Yang}
\author[addr1]{Jie Ruan}
\author[addr1]{Zhaocheng Li}
\author[addr2]{Erik Goodman}


\address[addr1]{Department of Electronic Engineering, Shantou University, Guangdong, China}

\address[addr2]{BEACON Center for the Study of Evolution in Action, Michigan State University. East Lansing, Michigan, USA.}

\begin{abstract}
This paper proposes a push and pull search method in the framework of differential evolution (PPS-DE) to solve constrained single-objective optimization problems (CSOPs). More specifically, two sub-populations, including the top and bottom sub-populations, are collaborated with each other to search global optimal solutions efficiently. The top sub-population adopts the pull and pull search (PPS) mechanism to deal with constraints, while the bottom sub-population use the superiority of feasible solutions (SF) technique to deal with constraints. In the top sub-population, the search process is divided into two different stages --- push and pull stages. In the push stage, a CSOP is optimized without considering any constraints, which can help to get across to infeasible regions. In the pull stage, the CSOP is optimized with an improved epsilon constraint-handling method, which can help the population to search for feasible solutions. An adaptive DE variant with three trial vector generation strategies --- DE /rand/1, DE/current-to-rand/1, and DE/current-to-pbest/1 is employed in the proposed PPS-DE. In the top sub-population, all the three trial vector generation strategies are used to generate offsprings, just like in CoDE. In the bottom sub-population, a strategy adaptation, in which the trial vector generation strategies are periodically self-adapted by learning from their experiences in generating promising solutions in the top sub-population, is used to choose a suitable trial vector generation strategy to generate one offspring. Furthermore, a parameter adaptation strategy from LSHADE44 is employed in both sup-populations to generate scale factor $F$ and crossover rate $CR$ for each trial vector generation strategy. Twenty-eight CSOPs with 10-, 30-, and 50-dimensional decision variables provided in the CEC2018 competition on real parameter single objective optimization are optimized by the proposed PPS-DE. The experimental results demonstrate that the proposed PPS-DE has the best performance compared with the other seven state-of-the-art algorithms, including AGA-PPS, LSHADE44, LSHADE44+IDE, UDE, IUDE, $\epsilon$MAg-ES and C$^2$oDE.
\end{abstract}


\end{frontmatter}


\section{Introduction}
\label{sec:int}

Many real-world optimization problems can be formulated as constrained optimization problems which have a set of constraints \cite{Floudas1990A}, \cite{gen2000genetic}. Without lose of generality, a constrained single-objective optimization problem (CSOP) can be defined as follows:
\begin{equation}
	\label{equ:csop_definition}
	\begin{cases}
		\mbox{minimize}   & f(\mathbf{x}) \\
		\mbox{subject to} & g_i(\mathbf{x}) \ge 0, i = 1,\ldots,q  \\
		                  & h_j(\mathbf{x}) = 0, j= 1,\ldots,p     \\
		                  & \mathbf{x} = (x_1,...,x_D) \in {S^D}
	\end{cases}
\end{equation}
where $f(\mathbf{x})$ is the objective function. $\mathbf{x}$ is the decision vector. $x_i$ is the $i$-th component of $\mathbf{x}$. $S=\prod_{i=1}^{D} {[L_i,\ U_i]}$ is the decision space, where $L_i$ and $U_i$ are the lower and the upper bounds of $x_i$. ${g_i}(\mathbf{x})$ denotes the $i$-th inequality constraint, and ${h_j}(\mathbf{x})$ denotes the $j$-th equality constraint.

In order to evaluate the constraint violation of a solution $\mathbf{x}$, the overall constraint violation method is a widely used method which summaries all the constraints into a scalar value $\phi(\mathbf{x})$ as follows:
\begin{eqnarray}
\label{equ:constraint}
\phi(\mathbf{x}) = \sum_{i=1}^{q} \max(g_i(\mathbf{x}),0) + \sum_{j = 1}^{p} \max(|h_j(\mathbf{x})|-\sigma, 0)
\end{eqnarray}
$\sigma$ is an extremely small positive number, which is set to $0.0001$ as suggested in Ref. \cite{wu2016problem}. If $\phi(\mathbf{x}) = 0$, $\mathbf{x}$ is a feasible solution, otherwise it is an infeasible solution.

As a kind of population-based optimization algorithms, evolutionary algorithms (EAs) have attracted lots of interest in solving CSOPs. Because they have not any requirements for the objectives and constraints of CSOPs. To solve CSOPs, there are two basic components in constrained EAs. One is the single-objective evolutionary algorithm (SOEA), and the other is the constraint-handling technique.

In terms of SOEAs, differential evolution (DE) is arguably one of the most powerful and versatile evolutionary optimizers in recent times \cite{5601760,DAS20161}. There are two main reasons. The first reason is that the structure of DE is very simple. It is very easy to implement a DE algorithm by using any currently popular programming languages. The second reason is that the number of parameters in DE is few. It is very convenient for a novice to solve optimization problems by using DE \cite{8315135}. In recent years, many different DE variants have been suggested, which include FADE \cite{liu2005fuzzy}, jDE \cite{4016057}, JADE \cite{5208221}, CoDE \cite{5688232}, SHADE \cite{tanabe2013success}, LSHADE \cite{6900380}, LSHADE44 \cite{7969504} and so on. In FADE \cite{liu2005fuzzy}, fuzzy logic controllers are employed to adapt the search parameters for the mutation operation and crossover operation. In jDE \cite{4016057}, a self-adaptive method is proposed to determine the values of the scale factor $F$ and the crossover rate $CR$. In JADE \cite{5208221}, a current-to-$p$best/1 with an optional external archive and a greedy mutation operator are proposed. It adaptively updates the control parameters $F$ and $CR$ in each generation. CoDE \cite{5688232} combines three trial vector generation strategies and three control parameter settings randomly to generate trial vectors. In SHADE \cite{tanabe2013success}, an adaptive technique of parameter settings by using successful historic memories is proposed to generate trial vectors. LSHADE \cite{6900380} is an improved version of SHADE, which reduces the population size linearly during the evolutionary process. As an variant of LSHADE, LSHADE44 \cite{7969504} proposes a strategy to select four different kinds of trial vector generation strategies adaptively.

The constraint-handling technique is the other key component in constrained EAs. Many constraint-handling methods have been proposed in evolutionary optimization \cite{MEZURAMONTES2011173,Coello:2017}. They can be generally classified into four different types, including penalty function methods, separation of objectives and constraints, multi-objective evolutionary algorithms (MOEAs) and hybrid methods \cite{MEZURAMONTES2011173,Coello:2017,CoelloCoello20021245}.

The penalty function method is a widely used method due to its simplicity in the constraint handling (\cite{Runarsson:2005jd}). It adopts a penalty factor $\lambda$ to maintain a balance between minimizing the objectives and satisfying the constraints. A CSOP is converted into an unconstrained single-objective optimization problem (SOP) by adding the overall constraint violation multiplied by a predefined penalty factor $\lambda$ to the objective \cite{CoelloCoello20021245}. If $\lambda = \infty$, this penalty function method is called a death penalty approach \cite{bdack1991survey}, which means that infeasible solutions are completely unacceptable. If $\lambda$ is a static value during the evolutionary process, it is called a static penalty method \cite{homaifar1994constrained}. If $\lambda$ is changing during the evolutionary process, it is called a dynamic penalty method \cite{joines1994use}. In the case in which $\lambda$ is dynamically changing according to the information collected during the evolutionary process, it is called an adaptive penalty approach \cite{bean1993dual,coit1996adaptive,ben1997genetic,4799193}. However, given an arbitrary CSOP, the ideal penalty factors cannot be known in advance. In fact, the ideal penalty factors should be dynamic parameters.

In the separation of objectives and constraints methods, the objectives and constraints are compared separately. Compared with the penalty function methods, there is no need to tune the penalty factors. This type of constraint-handling method has a relatively high impact in evolutionary optimization in recent years. Representative examples include the superiority of feasible (SF) solutions \cite{DEB2000311}, $\varepsilon$ constraint-handling method \cite{takahama2005constrained}, stochastic ranking approach (SR) \cite{runarsson2000stochastic}, and so on. In SF \cite{DEB2000311}, three basic rules are used to compare any two solutions. In rule 1, for two infeasible solutions, the one with less overall constraint violation is better. In rule 2, if one solution is feasible and the other is infeasible, the feasible one is preferred. In rule 3, for two feasible solutions, the one with a smaller objective value is better. In $\varepsilon$ constraint-handling method, the relaxation of the constraints is controlled by the epsilon level $\varepsilon$, which can help to maintain a search balance between feasible and infeasible regions during the evolutionary process. In the $\varepsilon$ constraint-handling method, if the overall constraint violation of a solution is less than $\varepsilon$, this solution is deemed feasible. Therefore, the epsilon level $\varepsilon$ is a critical parameter. In the case of $\varepsilon = 0$, $\varepsilon$ constraint-handling method is the same as SF \cite{996017}. Although $\varepsilon$ constraint-handling is a very popular method, controlling the $\varepsilon$ level properly is not at all trivial.

In SR \cite{runarsson2000stochastic}, a probability parameter $p_f \in [0,1]$ is employed to decide whether the comparison is based on objectives or constraints. For any two solutions, if a random number is less than or equal to $p_f$, the one with the smaller objective value is better---i.e., the comparison is based on objectives. If the random number is greater than $p_f$, the comparison is based on the overall constraint violation.

In order to balance the constraints and the objectives, some researchers adopt multi-objective evolutionary algorithms (MOEAs) to deal with constraints \cite{MezuraMontes:2011cj}. For example, the constraints of a CSOP can be converted into one or $k$ extra objectives. Then the CSOP is transformed into an unconstrained $2$- or $(1 + k)$-objective optimization problem, which can be solved by MOEAs. Representative examples include Cai and Wang's Method (CW) \cite{cai2006multiobjective} and the infeasibility driven evolutionary algorithms (IDEA) \cite{ray2009infeasibility}.

In the type of hybrid constraint-handling methods, several constraint-handling mechanisms are hybrid to deal with constraints. For example, the adaptive trade-off model (ATM) \cite{wang2008adaptive} uses two different constraint-handling mechanisms, including a multi-objective approach and an adaptive penalty function method, in different evolutionary stages. In Ref. \cite{qu2011constrained}, three different constraint-handling techniques, including $\varepsilon$ constraint-handling \cite{takahama2005constrained}, self-adaptive penalty functions (SP) \cite{4799193} and SF \cite{996017}, are employed to deal with constraints.

It can be concluded that most above-mentioned constraint-handling methods have some limitations. The recently proposed push and pull search \cite{FAN2018} is a general framework to deal with constrained optimization. It has already been proved that PPS has many advantages in deal with constrained multi-objective optimization problems \cite{FAN2018}. In this paper, we try to investigate the performance of PPS in solving CSOPs. The PPS method is integrated into an adaptive DE framework to solve CSOPs. The contributions of this paper are summarized as follows:
\begin{itemize}
  \item The push and pull search technique and SF constraint-handling method are successfully embedded into an adaptive DE framework for constrained single-objective optimization.
  \item Two sub-populations, which use different constraint-handling mechanisms and trial vector generation strategies, are collaborated with each other efficiently to search for global optimal solutions.
  \item The comprehensive experimental results indicate that the proposed PPS-DE provides state-of-the-art performance on the 28 CSOPs suggested in the CEC2018 competition on real parameter single objective optimization.
\end{itemize}

The rest of this paper is organized as follows. Section \ref{sec:related_work} introduces some related work adaptive DE algorithms. Section \ref{sec:proposed_method} introduces the proposed method PPS-DE in detail. Comprehensive experimental results and discussions are given in Section \ref{sec:exp_study}. Finally, conclusions are made in Section \ref{sec:con}.

\section{Related Work}
\label{sec:related_work}
In this section, some related work on adaptive DE algorithms are introduced. The proposed PPS-DE employs an adaptive DE algorithm which is inspired from some adaptive DE variants, including CoDE \cite{5688232}, C$^2$oDE \cite{8315135}, LSHADE44+IDE \cite{7969472}, UDE \cite{7969446}, IUDE \cite{Anupam2018} and AGA-PPS \cite{inbook}. The description of each DE algorithm is given as follows:

\begin{itemize}
  \item CoDE --- CoDE \cite{5688232} randomly combines three trial vector generation strategies and three control parameter settings to generate trial vectors. In CoDE \cite{5688232}, the strategy pool consists of DE/rand/1/bin, DE/rand/2/bin, and DE/current-to-rand/1 trial vector generation strategies. The parameter pool consists of three different control parameter settings, including [$F = 1.0, C_r = 0.1$], [$F = 1.0, C_r = 0.9$] and [$F = 0.8, C_r = 0.2$]. When generating offsprings, three trial vectors are created by using the three trial vector generation strategies with randomly selected control parameter settings from the parameter pool. Then, the best trial vector is selected to update its parent. The experimental results demonstrated that the overall performance of CoDE is better than four other state-of-the-art DE variants, i.e., JADE \cite{5208221}, jDE \cite{4016057}, SaDE \cite{4632146}, and EPSDE \cite{MALLIPEDDI20111679}, and three non-DE variants, i.e., CLPSO \cite{1637688}, CMA-ES \cite{CMA-ES2001}, and GL-25 \cite{GARCIAMARTINEZ20081088} on 25 global numerical optimization problems used in the CEC2005 special session on real-parameter optimization.

  \item C$^2$oDE --- C$^2$oDE \cite{8315135} is an extension of CoDE \cite{5688232} for solving CSOPs. It also adopts three different trial vector generation strategies, including DE/current-to-rand/1, DE/current-to-best/1, and modified DE/rand-to-best/1, to balance diversity and convergence of a working population. In terms of constraint-handling, a new comparison rule, which combines the feasibility rule \cite{DEB2000311} with the $\varepsilon$ constrained method \cite{takahama2005constrained}, is proposed. When generating offsprings, three trial vectors are created by using the three trial vector generation strategies. Then, the best trial vector is selected by using the feasibility rule, and the selected trial vector is used to update its parent by using the $\varepsilon$ constrained method. Moreover, a restart scheme is proposed to help the population jump out of a local optimal in the infeasible region for some extremely complex CSOPs.

  \item LSHADE44 --- LSHADE44 \cite{7969504} is an enhanced version of LSHADE \cite{6900380} which was a first ranked algorithm at the CEC2014 competition on real-Parameter single objective optimization. In LSHADE44, four different trial vector generation strategies, including DE/current-to-pbest/1/bin, DE/current-to-pbest/1/exp, DE/rand1/1/bin, and DE/randr1/1/exp, are adopted to generate an offspring. The newly generated offspring updates its parent by using the feasibility rule \cite{DEB2000311} to deal with constraints.

  \item LSHADE44+IDE --- The search process of LSHADE44+IDE \cite{7969472} is divided into two stages. In the first stage, the search of feasible individuals is carried out by minimization of the mean constraint violation. When a number of feasible individuals given a priori is found or the predefined portion of function evaluations (FES) is consumed, the search process is switched to the second stage. In the second stage, the function value is minimized until the stopping criteria is met. If a sufficient amount of feasible individuals is found in the first stage, the feasible solutions are adopted as an initial population for the second search stage, otherwise the individuals with the smallest mean constraint violation are used as an initial population for the second stage. An adaptive version of DE named LSHADE44 \cite{7744403} with reduction of the population size of four DE strategies is used in the first search stage. The adaptive DE variant with individual-dependent technique \cite{6913512} is employed in the second search stage.
  \item UDE --- UDE \cite{7969446} is inspired from some popular DE variants, including CoDE \cite{5688232}, JADE \cite{5208221}, SaDE \cite{4632146}, and ranking-based mutation operator \cite{6423878}. It ranked second in the CEC 2017 competition on constrained real parameter optimization. In UDE, three trial vector generation strategies and two types of control parameter settings are combined. More specifically, UDE divides the working population into two sub-populations. In the top sup-population, UDE used all the three trial vector generation strategies on each target vector, just like in CoDE \cite{5688232}. In the bottom sub-population, strategy adaptation is applied to select a trial vector generation strategy to generate a offspring. In the strategy adaptation, the three trial vector generation strategies are periodically self-adapted by learning from their experiences in generating promising solutions in the top sub-population. In addition, a DE mutation strategy based on local search operation is adopted in UDE. A static penalty method is used to deal with constraints in UDE.
  \item IUDE --- IUDE \cite{Anupam2018} is an improved version of UDE \cite{7969446}. It ranked first in the CEC2018 competition on real parameter single objective optimization. In IUDE, the constraint-handling method is a combination of $\varepsilon$ constraint-handling technique \cite{takahama2005constrained} and superiority of feasible solutions method \cite{DEB2000311}, while in UDE, only the static penalty method is adopted to deal with constraints. Furthermore, IUDE employs the parameter adaptation technique in LSHADE44 \cite{7969504} to generate offsprings, while UDE utilizes a control parameter pool to generate offsprings.
  \item AGA-PPS --- AGA-PPS \cite{inbook} adopted an adaptive method to select recombination operators, including differential evolution (DE) operators and polynomial operators. Moreover, a push and pull search (PPS) method is employed to deal with constraints. The PPS has two search stages --- the push stage and the pull stage. In the push stage, a CSOP is optimized without considering constraints. In the pull stage, the CSOP is optimized with an improved epsilon constraint-handling method. The experimental results show that AGA-PPS is significantly better than other three DEs (LSHADE44+IDE, LSHADE44 and UDE) on the CEC 2017 competition on constrained real parameter optimization, which manifests that AGA-PPS is a quite competitive algorithm for solving CSOPs.
\end{itemize}

\section{Proposed Method}
\label{sec:proposed_method}
In this section, the proposed PPS-DE algorithm is presented. The proposed PPS-DE is a significantly enhanced version of AGA-PPS \cite{inbook}. The primary feature of the proposed PPS-DE lies in strengthening the DE algorithm and the constraint-handling method. PPS-DE is inspired from the following state-of-the-art DE variants, including CoDE \cite{5688232}, C$^2$oDE \cite{8315135}, LSHADE44+IDE \cite{7969472}, UDE \cite{7969446}, IUDE \cite{Anupam2018} and AGA-PPS \cite{inbook}. PPS-DE uses three different trial vector generation strategies, including modified DE/rand/1/bin, DE/current-to-pbest/1, and DE/current-to-rand/1, to generate three trial vectors. In PPS-DE, the working population is divided into two sub-populations, including the top and the bottom sub-populations. In the top sub-population, PPS-DE employs all the three trial vector generation strategies on each target vector, just like in CoDE \cite{5688232} and C$^2$oDE \cite{8315135}. In the bottom sub-population, an strategy adaptation, in which the trial vector generation strategies are periodically adapted by learning from their experiences in generating successful solutions in the top sub-population, is employed to select a trial vector generation strategy to generate one trial vector. The constraint-handling in the top sub-population is based on the PPS, and the bottom sub-population adopts the feasibility rule \cite{DEB2000311} to deal with constraints. Furthermore, the control parameter settings adaptation strategy proposed in LSHADE44 \cite{7969504} is also used in the PPS-DE algorithm. In the replacement process, the PPS is used to select individuals into the next generation.

\subsection{Push and Pull Search}
Push and pull search (PPS) is a general framework which is aim to solve constrained optimization problems \cite{FAN2018}. It first proposed to solve constrained multi-objective optimization problems (CMOPs), which is able to balance objective minimization and constraint satisfaction. The PPS divides the search process into two different stages. In the first stage, only the objectives are optimized, which means the working population is pushed toward the unconstrained global optimum without considering any constraints. In the pull stage, an improved epsilon constraint-handling approach is adopted to pull the working population to the constrained global optimum. In CMOPs, the influence of infeasible regions on Pareto fronts (PFs) can be classified into three different situations \cite{FAN2018}. In the first situation, infeasible regions block the way towards the PF. In the second situation, the unconstrained PF is covered by infeasible regions and all of it is infeasible. In the last situation, infeasible regions make the original unconstrained PF partially feasible.

In CSOPs, the influence of infeasible regions on the global optimum can be categorized into two different situations. In the first situation, infeasible regions block the way towards the global optimum, and the constrained global optimum is the same as its unconstrained global optimum. In the second situation, the unconstrained is covered by infeasible regions and the unconstrained global optimum is different to its constrained global optimum. For CSOPs, we also need to trade off objective minimization and constraint satisfaction. Therefore, it is quite natural to use PPS to solve CSOPs.

In the push search stage, a newly generated solution $\mathbf{y}^i$ is retained into the next generation based on the objective value as described in Algorithm \ref{alg:push}.
\begin{algorithm}
	\Fn{result = PushSearch($\mathbf{x}^i$,$\mathbf{y}^i$)}{
	$result = false$\\
	\If{$f(\mathbf{y}^i) \leq f(\mathbf{x}^i)$}{
	$\mathbf{x}^i$ = $\mathbf{y}^i$\\
	$result = true$
	}
	\Return $result$
	}
	\caption{Push Search}
	\label{alg:push}
\end{algorithm}

In the pull stage, infeasible solutions are pulled to the feasible regions by using the improved epsilon constraint-handling method. The details can be found in Ref. \cite{FAN2018}. A newly generated solution $\mathbf{y}^i$ is selected for survival into the next generation based on the objective value, the overall constraint violation $\phi(\mathbf{y}^i)$ and the value of $\varepsilon(k)$, as illustrated by Algorithm \ref{alg:pull}.

\begin{algorithm}
	\Fn{result = PullSearch($\mathbf{x}^i$,$\mathbf{y}^i$,$\varepsilon(k)$)}{
	$result = false$

	\uIf{$\phi(\mathbf{y}^i) \le \varepsilon(k)$ and $\phi(\mathbf{x}^i) \le \varepsilon(k)$}{
	\If{$f(\mathbf{y}^i) \leq f(\mathbf{x}^i)$}{
	$\mathbf{x}^j$ = $\mathbf{y}^i$;
	$result = true$
	}
	}
	\uElseIf{$\phi(\mathbf{y}^i) == \phi(\mathbf{x}^i)$}{
	\If{$f(\mathbf{y}^i) \leq f(\mathbf{x}^i)$}{
	$\mathbf{x}^i$ = $\mathbf{y}^i$;
	$result = true$
	}
	}
	\ElseIf{$\phi(\mathbf{y}^i) < \phi(\mathbf{x}^i)$}{
		$\mathbf{x}^i$ = $\mathbf{y}^i$;
		$result = true$
	}
	\Return $result$
	}
	\caption{Pull Search}
	\label{alg:pull}
\end{algorithm}

When solving CSOPs, the decision as to when to switch from the push to the pull search process is also very critical in the PPS. A strategy for when to switch the search behavior is suggested as follows.
\begin{eqnarray}
	\label{equ:r_rate}
	r_G \equiv \frac{f(\mathbf{x}^j_{G-L}) - f(\mathbf{x}^i_{G})}{ \max \{|f(\mathbf{x}^j_{G-L})|,\Delta\}} \le \epsilon
\end{eqnarray}
where $r_G$ represents the change rate of the minimal objective value during the last $L$ generations. $i=\arg_i \min f(\mathbf{x}^i_{G})$ and $j=\arg_j \min f(\mathbf{x}^j_{G-L})$ are the indexes of solutions with the minimum objective values in generation $G$ and $G-L$, respectively. $\epsilon$ is a user-defined parameter. In this paper, we have set $\epsilon = 1e-3$. $\Delta$ is a very small positive number, which is used to make sure that the denominator in Eq. \eqref{equ:r_rate} is not equal to zero. In this paper, $\Delta$ is set to $1e-6$. At the beginning of the search, $r_G$ is initialized to $1.0$. At each generation, $r_G$ is updated according to Eq. \eqref{equ:r_rate}. If $r_k$ is less than or equal to the predefined threshold $\epsilon$, the search behavior is switched to the pull search.

\subsection{Trial Vector Generation Strategy Adaptation}
As discussed above, the working population of PPS-DE is divided into two sub-populations, including the top sub-population $P_t$ and the bottom sub-population $P_b$. The top sub-population $P_t$ adopts three different trial vector generation strategies, including modified DE/rand/1/bin, DE/current-to-pbest/1, and DE/current-to-rand/1, to generate three trial vectors. The trial vector generation strategy with the best trial vector scores a win according to the PPS method. At each generation, the success rate of each trial vector generation strategy is calculated over the previous $L_p$ generations. For example, $NW_1, NW_2$, and $NW_3$ are the number of wins of trial vector generation strategy 1, 2, and 3 over the previous $L_p$ generations. The success rate $SR_i$ of trial vector generation strategy $i$ is defined as $SR_i = NW_i / (NW_1 + NW_2 + NW_3)$.

In the bottom subpopulation $P_b$, a trial vector generation strategy is selected according to its success rate which is calculated in the top sub-population $P_t$. Then, the selected trial vector generation strategy is employed to generate a trial vector. It is worth noting that each trial vector generation strategy has the same probability to be selected when the generation counter $k$ is less than $L_p$.

\subsection{Control Parameter Settings Adaptation}
In PPS-DE, the parameter adaptation principle of LSHADE44 \cite{7969504} is used in both sub-populations. In PPS-DE, three trial vector generation strategies are employed. Each trial vector generation strategy needs to set two parameters, including the scale factor $F$ and the crossover rate $CR$. Three pairs of memories $M_F$ and $M_{CR}$ for adaptation of $F$ and $CR$ are employed in PPS-DE. For each strategy, PPS-DE stores successful values of parameters $F$ and $CR$ into separate sets $S_F$ and $S_{CR}$ during a generation. Then, all three pairs of memories $M_F$ and $M_{CR}$ are adapted according to the values in sets $S_F$ and $S_{CR}$. At the beginning of each generation, all sets $S_F$ and $S_{CR}$ are reset to empty sets $\emptyset$. Each $k$ (There are three pointers $k$) is set to $k = 1$ at the beginning of the search. If there is a change in a pair of memory, $k$ is increased by 1. If $k > H$, where $H$ is the size of each historical memory, $k$ is reset to 1. At the beginning of the search, $M_{F_k}$ and $M_{CR_k}$ are initialized to 0.5. After each generation $M_{F_k}$ and $M_{CR_k}$ are calculated as follows.

 \begin{equation}\label{equ:mfk}
   M_{F_k}  = mean_{WL}(S_F) \ \ \ \  \text{if}\ \ \ \ S_F \neq \emptyset
 \end{equation}

 \begin{equation}\label{equ:mcrk}
   M_{CR_k}  = mean_{WA}(S_{CR}) \ \ \ \  \text{if}\ \ \ \ S_{CR} \neq \emptyset
 \end{equation}

 \begin{equation}\label{equ:meanwl}
   mean_{WL}(S_F) = \frac{\sum_{t = 1}^{|S_F|} w_{t} S_F^2(t)}{\sum_{t = 1}^{|S_F|} w_{t} S_F(t)}
 \end{equation}

  \begin{equation}\label{equ:meanwa}
   mean_{WA}(S_{CR}) = \sum_{t = 1}^{|S_{CR}|} w_{t} S_{CR}(t)
 \end{equation}

 \begin{equation}\label{wt}
   w_t = \frac{\Delta func_t}{\sum_{u = 1}^{|S_{CR}|}\Delta func_u}
 \end{equation}

\begin{equation}\label{equ:funct}
   \Delta func_t = |func(x_t) - func(y_t)|
\end{equation}
where $func$ is objective function $f$ in the case of the old point $x_t$ is replaced by $y_t$ because $f(y_t)$ was less or equal than $f(x_t)$. In the case of the old point $x_t$ is replaced by $y_t$ because $\phi(y_t)$ was less or equal than $\phi(x_t)$, $func$ is overall constraint violation function $\phi$.

The scale factor $F$ and the crossover rate $CR$ are generated as follows. In the set $\{1,\ldots,H\}$, a rand number $k$ is selected first. Then, $F$ is a random number from the Cauchy distribution with parameters $(M_{F_k},0.1)$. $F$ is regenerated until it is bigger than 0. If $F > 1$, $F$ is set to one. $CR$ is a random number from the Gauss distribution with parameters $(M_{CR_k},0.1)$. $CR$ is truncated into interval [0,1].

\subsection{Constraint Handling}
In PPS-DE, two different kinds of constraint handling methods are employed. They are PPS technique \cite{FAN2018} and the superiority of feasible (SF) solutions method \cite{DEB2000311}. More specifically, in the top subpopulation, the PPS technique is adopted as the constraint-handling method to select the best trial vectors. The SF constraint-handling method is used to sort each solution of a population in increasing order. Then the sorted population is divided into the top and the bottom sub-populations. In the replacement process, the PPS technique is also used to select solutions into the next generation.

\subsection{The Framework of the Proposed Method}
Algorithm \ref{alg:pps-de} outlines the pseudocode of the proposed PPS-DE algorithm. The generation counter $G$ and the population $P$ are initialized at line 1. At line 2, the initialized population is evaluated, and the number of consumed function evaluations is recorded.  The number of wins of each trial vector generation strategy $i$ is initialized at line 3. Memories $M_{F_k}^j$ and $M_{CR_k}^j$ which are used to set $F$ and $CR$ are also initialized at line 3.

The algorithm repeats lines 4-24 until $FES$ is greater than $MaxFES$. At line 5, $\varepsilon(G)$ is calculated according to the PPS Method. The working population $P$ is divided into top and bottom sub-populations at line 6. Lines 7-11 show the process of generating offsprings in the top sub-population. At line 12, the best $T$ offsprings are selected out from the newly generated solutions according to the PPS method. The number of wins corresponding to winning trial vector generation strategy is updated at line 13. The success rate $SR_j$ of trial vector generation strategy $j$ is calculated at line 14. Lines 15-19 show the process of generating offsprings in the bottom sub-population. It is worth noting that only one trial vector is generated at each iteration. At line 20, the one-to-one comparison is employed to select solutions to the next generation according to the PPS method. Three pairs of memories $M_{F_k}^j$ and $M_{CR_k}^j$ for adaptation of $F$ and $CR$ are updated at line 21. The best solution in the current generation is selected out at line 22. Finally, the generation counter $G$ is updated at line 23.

\begin{algorithm}
	\KwIn{$N_{p}$: the population size. $MaxFEs$: the max number of function evaluation. $T=0.5N_p$: the size of top sub-population. $H$: the length of the historic memory.}
	\KwOut{$B_{fs}:$ the best feasible solution.}

        Set $G = 0$; Generate a population $P=\{\mathbf{x}_1,...,\mathbf{x}_{N}\}$.\\
		Evaluate the objective values $f(x_{i})$ and the overall constraint violation $\phi(x_{i})$; Set $FES=N_{p}$.\\
		Initialize $NW_{j}=0$, $M_{F_k}^j= 0.5$ and $M_{CR_k}^j = 0.5$, for $j=1,2,3$ and $k = 1,2, \ldots, H$.\\
	\While{$FES \leq MaxFES$}{
           Calculate $\varepsilon(G)$.\\
           Divide $P$ into top sub-population $P_t$ and bottom sub-population $P_b$ according to the SF constraint-handling method.\\
           	\For{$i \leftarrow 1$ \KwTo $T$}{
            For each target vector $x_{i,G} \in P_t$, use three DE strategies to generate three trial vectors $u_{i1,G}$, $u_{i2,G}$,  $u_{i3,G}$;\\
             Evaluate the fitness function values of the three trial vectors $u_{i1,G}$, $u_{i2,G}$, $u_{i3,G}$.\\
		   $FES=FES+3$;\\
		   }
		   Use the PPS method to select best $T$ offsprings from the newly generated solutions.\\
		   Update the number of wins corresponding to winning trial vector generation strategy $j$ as $NW_{j} = NW_{j} + 1$;\\
		   Update $SR_{j}=NW_{j}/(NW_{1}+NW_{2}+NW_{3})$. \\

           \For{$i \leftarrow T + 1$ \KwTo $N_p$}{
           For each target vector $x_{k,G} \in P_b$, select a trial vector strategy according to $SR_{j}$ to generate one trial vectors $u_{k,G}$;\\
           Evaluate the fitness function value of the trial vectors $u_{k,G}$.\\
		   $FES=FES + 1$;	\\	
		   }

		   Using the one-to-one comparison to select solutions to the next generation to form population $P$ according to the PPS method. \\
           Updating $M_{F_k}^j$ and $M_{CR_k}^j$ according to Eq. \eqref{equ:mfk}-\eqref{equ:funct};\\
          Finding the best solution according to the SF constraint handling method in the current population $P$;\\
          $G = G + 1$;	\\
	}
	\caption{PPS-DE}
	\label{alg:pps-de}
\end{algorithm}

\section{Experimental Study}
\label{sec:exp_study}

\subsection{Experimental settings}
Seven state-of-the art constrained EAs, including AGA-PPS \cite{inbook}, LSHADE44 \cite{7969504}, LSHADE44+IDE \cite{7969472}, UDE \cite{7969446}, IUDE \cite{Anupam2018}, $\epsilon$MAg-ES \cite{8477950} and C$^2$oDE \cite{8315135}, are employed to compare with the proposed PPS-DE on the 28 benchmark problems with 10-, 30- and 50-dimensional decision variables provided in the CEC2018 competition on real parameter single objective optimization. Each algorithm runs for 25 times independently on the 28 test instances. The parameter settings of each algorithm are listed as follows:
\begin{enumerate}
\item[1)]  Population size: $N_{P}=5D$, where $D$ is dimension of problem.
\item[2)]  Size of top sub-population $T=0.5N$.
\item[3)]  Learning period $L=25$ generations.
\item[4)]  DE/current-to-$p$best/1 parameter: $p=5$.
\item[5)]  The max number of function evaluation: $MaxFEs=20000D$
\end{enumerate}

Friedman aligned test is used to check whether the difference between the proposed PPS-DE and the compared algorithms is statistically significant. The Friedman aligned test is carried out with a 0.05 significance level.

\subsection{Discussion of Experiments}
The mean values and the standard deviations of the objectives on the test instances C01 - C28 with $D$ = 10 achieved by eight algorithms in 25 independent runs are listed in Table \ref{tab:10d}. The Friedman aligned test indicates that PPS-DE ranks the highest among the eight algorithms, as shown in the last row of Table \ref{tab:10d}. The p-value computed through the statistics of the Friedman aligned test is 0, which strongly suggests the existence of significant differences among the eight tested algorithms. For C01-06, C13, C16, C19, C25 and C28 with 10-dimensional decision vectors, PPS-DE can achieved the global optimal solutions steadily. In the 28 test instances, PPS-DE has the best performance on 16 test problems among the eight tested algorithms, which indicates the superiority of the PPS-DE.

The statistic results of the objectives on the test instances C01 - C28 with $D$ = 30 achieved by eight algorithms in 25 independent runs are listed in Table \ref{tab:30d}. On the 28 test instances, PPS-DE has the best performance on 10 test problems among the eight tested algorithms. The Friedman aligned test also indicates that PPS-DE ranks first among the eight algorithms, as shown in the last row of Table \ref{tab:30d}. The p-value computed through the statistics of the Friedman aligned test is 0, which strongly suggests the existence of significant differences among the eight tested algorithms.

Table \ref{tab:50d} shows the mean values and the standard deviations of the objectives on the test instances C01 - C28 with $D$ = 50 achieved by eight algorithms in 25 independent runs. In the 28 test instances, PPS-DE has the best performance on 10 test problems among the eight tested algorithms. The Friedman aligned test also indicates that PPS-DE ranks the highest among the eight algorithms, as shown in the last row of Table \ref{tab:50d}. The p-value computed through the statistics of the Friedman aligned test is 0, which strongly suggests the existence of significant differences among the eight tested algorithms.

From the above observation, it is clear that PPS-DE is significantly better than the other seven algorithms on most of the 28 test instances. One possible reason is that, among the 28 test problems, there are many instances whose global optimal solutions are the same as those of their unconstrained counterparts. In PPS-DE, the global optimal solutions can be achieved in the push stage without dealing with any constraints.

\begin{table*}[htbp]
  \centering
  \caption{The mean value and the standard deviation of the objective during the 25 runs on the test instances C01 - C28 with $D$ = 10.}
  \label{tab:10d}
  \scalebox{0.7}[0.7]{
    \begin{tabular}{cc|c|c|c|c|c|c|c|c}
    \toprule
    \multicolumn{2}{c|}{Test Instances} & \multicolumn{1}{c|}{AGA-PPS} & \multicolumn{1}{c|}{LSHADE44+IDE} & \multicolumn{1}{c|}{LSHADE44} & \multicolumn{1}{c|}{UDE} & \multicolumn{1}{c|}{IUDE} & \multicolumn{1}{c|}{$\epsilon$MAg-ES} & \multicolumn{1}{c|}{C$^2$oDE} & \multicolumn{1}{c}{PPS-DE} \\
    \hline

     \multirow{2}[0]{*}{C01} & mean  & \textbf{0.00E+00} & \textbf{0.00E+00} & \textbf{0.00E+00} & 5.03E-15 & \textbf{0.00E+00} & 1.65E-30 & \textbf{0.00E+00} & \textbf{0.00E+00} \\
          & std   & 0.00E+00 & 0.00E+00 & 0.00E+00 & 5.28E-15 & 0.00E+00 & 7.58E-30 & 0.00E+00 & 0.00E+00 \\
    \multirow{2}[0]{*}{C02} & mean  & 1.01E-30 & \textbf{0.00E+00} & \textbf{0.00E+00} & 6.44E-15 & \textbf{0.00E+00} & \textbf{0.00E+00} & \textbf{0.00E+00} & \textbf{0.00E+00} \\
          & std   & 3.50E-30 & 0.00E+00 & 0.00E+00 & 8.16E-15 & 0.00E+00 & 0.00E+00 & 0.00E+00 & 0.00E+00 \\
    \multirow{2}[0]{*}{C03} & mean  & 7.58E+00 & 3.26E+05 & 3.15E+04 & 7.74E+01 & 3.54E+01 & 4.73E-31 & \textbf{0.00E+00} & \textbf{0.00E+00} \\
          & std   & 2.63E+01 & 2.58E+05 & 3.70E+04 & 8.30E+00 & 3.77E+01 & 1.73E-30 & 0.00E+00 & 0.00E+00 \\
    \multirow{2}[0]{*}{C04} & mean  & 1.63E+00 & 1.44E+01 & 1.36E+01 & 2.51E+01 & 2.90E+00 & 2.98E+01 & 1.36E+01 & \textbf{0.00E+00} \\
          & std   & 4.50E+00 & 1.15E+00 & 6.15E-02 & 9.10E+00 & 5.95E+00 & 1.76E+01 & 2.74E-07 & 0.00E+00 \\
    \multirow{2}[0]{*}{C05} & mean  & \textbf{0.00E+00} & \textbf{0.00E+00} & \textbf{0.00E+00} & 1.68E+00 & 1.74E-30 & \textbf{0.00E+00} & \textbf{0.00E+00} & \textbf{0.00E+00} \\
          & std   & 0.00E+00 & 0.00E+00 & 0.00E+00 & 9.72E-01 & 6.02E-30 & 0.00E+00 & 0.00E+00 & 0.00E+00 \\
    \multirow{2}[0]{*}{C06} & mean  & \textbf{0.00E+00} & 8.08E+02 & 6.49E+02 & 8.71E+01 & \textbf{0.00E+00} & 3.58E+01 & \textbf{0.00E+00} & \textbf{0.00E+00} \\
          & std   & 0.00E+00 & 5.45E+02 & 2.84E+02 & 3.18E+01 & 0.00E+00 & 3.82E+01 & 0.00E+00 & 0.00E+00 \\
    \multirow{2}[0]{*}{C07} & mean  & 1.36E+02 & -3.40E+01 & 3.74E+00 & -6.46E+00 & -2.77E+02 & -3.17E+02 & -2.88E+02 & \textbf{-3.57E+02} \\
          & std   & 6.86E+01 & 5.70E+01 & 6.96E+01 & 9.55E+01 & 1.10E+02 & 8.32E+01 & 9.25E+01 & 1.45E+02 \\
    \multirow{2}[0]{*}{C08} & mean  & 1.35E-03 & 0.00E+00 & \textbf{-1.35E-03} & -1.34E-03 & \textbf{-1.35E-03} & -1.35E-03 & -1.35E-03 & \textbf{-1.35E-03} \\
          & std   & 4.43E-19 & 0.00E+00 & 2.21E-19 & 8.33E-06 & 0.00E+00 & 0.00E+00 & 4.05E-13 & 0.00E+00 \\
    \multirow{2}[0]{*}{C09} & mean  & 4.98E-03 & 0.00E+00 & -4.97E-03 & -4.98E-03 & \textbf{-4.98E-03} & -4.98E-03 & -4.98E-03 & -4.98E-03 \\
          & std   & 2.66E-18 & 0.00E+00 & 2.44E-05 & 1.08E-10 & 0.00E+00 & 0.00E+00 & 0.00E+00 & 0.00E+00 \\
    \multirow{2}[0]{*}{C10} & mean  & 5.10E-04 & 0.00E+00 & \textbf{-5.10E-04} & -5.08E-04 & \textbf{-5.10E-04} & -5.10E-04 & -5.10E-04 & \textbf{-5.10E-04} \\
          & std   & 2.21E-19 & 0.00E+00 & 1.11E-19 & 2.02E-06 & 7.11E-16 & 0.00E+00 & 3.50E-13 & 1.78E-15 \\
    \multirow{2}[0]{*}{C11} & mean  & 1.69E-01 & 0.00E+00 & -1.69E-01 & -6.00E+00 & -8.01E-01 & -1.68E-01 & -1.69E-01 & \textbf{-4.60E+01} \\
          & std   & 1.05E-03 & 0.00E+00 & 2.83E-17 & 1.00E-04 & 1.76E-06 & 5.13E-03 & 1.40E-09 & 1.59E+02 \\
    \multirow{2}[0]{*}{C12} & mean  & 3.99E+00 & \textbf{3.99E+00} & 3.99E+00 & 3.99E+00 & 3.99E+00 & 7.00E+00 & 3.99E+00 & 4.01E+00 \\
          & std   & 3.07E-05 & 0.00E+00 & 2.32E-03 & 6.00E-06 & 1.23E-03 & 7.03E+00 & 5.79E-04 & 1.04E-01 \\
    \multirow{2}[0]{*}{C13} & mean  & 1.60E-01 & \textbf{0.00E+00} & \textbf{0.00E+00} & 1.11E+01 & \textbf{0.00E+00} & 1.59E-01 & \textbf{0.00E+00} & \textbf{0.00E+00} \\
          & std   & 7.97E-01 & 0.00E+00 & 0.00E+00 & 2.26E+01 & 0.00E+00 & 7.97E-01 & 0.00E+00 & 0.00E+00 \\
    \multirow{2}[0]{*}{C14} & mean  & \textbf{2.38E+00} & 3.00E+00 & 2.88E+00 & 2.74E+00 & 2.38E+00 & 2.87E+00 & 2.38E+00 & 2.38E+00 \\
          & std   & 9.07E-16 & 0.00E+00 & 2.04E-01 & 3.22E-01 & 1.36E-15 & 7.63E-01 & 1.36E-15 & 1.36E-15 \\
    \multirow{2}[0]{*}{C15} & mean  & \textbf{4.62E+00} & 1.13E+01 & 1.45E+01 & 6.75E+00 & 6.38E+00 & 7.61E+00 & 6.63E+00 & 6.13E+00 \\
          & std   & 1.93E+00 & 2.34E+00 & 3.77E+00 & 2.03E+00 & 4.11E+00 & 6.47E+00 & 3.83E+00 & 4.53E+00 \\
    \multirow{2}[0]{*}{C16} & mean  & \textbf{0.00E+00} & 4.04E+01 & 4.07E+01 & 6.28E+00 & \textbf{0.00E+00} & \textbf{0.00E+00} & \textbf{0.00E+00} & \textbf{0.00E+00} \\
          & std   & 0.00E+00 & 6.03E+00 & 6.72E+00 & 1.29E-04 & 0.00E+00 & 0.00E+00 & 0.00E+00 & 0.00E+00 \\
    \multirow{2}[0]{*}{C17} & mean  & 6.42E-01 & 1.00E+00 & 9.11E-01 & 1.05E+00 & \textbf{1.99E-02} & 7.35E-01 & NaN   & 2.74E-01 \\
          & std   & 5.83E-01 & 0.00E+00 & 1.77E-01 & 1.34E-01 & 4.48E-02 & 3.22E-01 & NaN   & 4.45E-01 \\
    \multirow{2}[0]{*}{C18} & mean  & 3.66E+01 & 3.17E+03 & 2.11E+03 & 2.36E+03 & \textbf{1.17E-01} & 3.66E+01 & NaN   & 3.72E+00 \\
          & std   & 1.26E-05 & 2.41E+03 & 2.11E+03 & 1.76E+03 & 1.31E+01 & 1.65E-05 & NaN   & 8.37E+00 \\
    \multirow{2}[0]{*}{C19} & mean  & \textbf{0.00E+00} & \textbf{0.00E+00} & 1.45E-06 & 2.72E-03 & \textbf{0.00E+00} & 1.12E+00 & NaN   & \textbf{0.00E+00} \\
          & std   & 0.00E+00 & 0.00E+00 & 3.12E-07 & 6.66E-03 & 0.00E+00 & 2.34E+00 & NaN   & 0.00E+00 \\
    \multirow{2}[0]{*}{C20} & mean  & 5.76E-01 & 4.16E-01 & \textbf{1.93E-01} & 1.65E+00 & 6.99E-01 & 1.17E+00 & 4.74E-01 & 5.83E-01 \\
          & std   & 1.63E-01 & 1.24E-01 & 5.79E-02 & 3.93E-01 & 1.16E-01 & 3.93E-01 & 1.34E-01 & 1.10E-01 \\
    \multirow{2}[0]{*}{C21} & mean  & 4.92E+00 & \textbf{3.99E+00} & 3.99E+00 & 6.24E+00 & \textbf{3.99E+00} & 4.41E+00 & 4.41E+00 & \textbf{3.99E+00} \\
          & std   & 3.22E+00 & 0.00E+00 & 5.20E-04 & 6.23E+00 & 4.12E-05 & 2.12E+00 & 2.12E+00 & 4.99E-03 \\
    \multirow{2}[0]{*}{C22} & mean  & 4.78E-01 & 1.60E-01 & 6.38E-01 & 1.25E+01 & 3.14E+00 & 6.38E-01 & \textbf{3.41E-27} & 3.66E-27 \\
          & std   & 1.32E+00 & 7.97E-01 & 1.49E+00 & 2.47E+01 & 1.24E+01 & 1.49E+00 & 7.17E-29 & 9.99E-28 \\
    \multirow{2}[0]{*}{C23} & mean  & 2.42E+00 & 3.02E+00 & 2.54E+00 & 2.75E+00 & \textbf{2.38E+00} & 2.50E+00 & 2.38E+00 & \textbf{2.38E+00} \\
          & std   & 9.45E-02 & 2.09E-01 & 2.49E-01 & 2.95E-01 & 6.65E-15 & 3.27E-01 & 0.00E+00 & 8.84E-16 \\
    \multirow{2}[0]{*}{C24} & mean  & \textbf{2.73E+00} & 8.77E+00 & 8.64E+00 & 6.00E+00 & 5.50E+00 & 6.13E+00 & 4.24E+00 & 4.99E+00 \\
          & std   & 1.04E+00 & 1.10E+00 & 9.07E-01 & 1.18E+00 & 3.01E+00 & 2.22E+00 & 2.03E+00 & 4.12E+00 \\
    \multirow{2}[0]{*}{C25} & mean  & \textbf{0.00E+00} & 3.77E+01 & 3.87E+01 & 6.35E+00 & \textbf{0.00E+00} & \textbf{0.00E+00} & \textbf{0.00E+00} & \textbf{0.00E+00} \\
          & std   & 0.00E+00 & 7.31E+00 & 5.19E+00 & 3.14E-01 & 0.00E+00 & 0.00E+00 & 0.00E+00 & 0.00E+00 \\
    \multirow{2}[0]{*}{C26} & mean  & 7.01E-01 & 9.37E-01 & 1.06E+00 & 1.02E+00 & \textbf{8.65E-02} & 7.54E-01 & NaN   & 3.02E-01 \\
          & std   & 4.71E-01 & 3.72E-01 & 3.23E-01 & 5.17E-02 & 1.84E-01 & 3.25E-01 & NaN   & 4.41E-01 \\
    \multirow{2}[0]{*}{C27} & mean  & \textbf{3.66E+01} & 7.94E+03 & 3.55E+03 & 6.72E+03 & 7.67E+01 & 7.59E+01 & NaN   & 6.40E+01 \\
          & std   & 2.00E-05 & 9.15E+03 & 4.81E+03 & 6.80E+03 & 4.75E+01 & 1.37E+02 & NaN   & 6.03E+01 \\
    \multirow{2}[0]{*}{C28} & mean  & 6.30E+00 & 1.08E+01 & 1.97E+01 & 9.76E+00 & 4.57E+00 & 8.68E+00 & NaN   & \textbf{0.00E+00} \\
          & std   & 6.55E+00 & 1.63E+01 & 1.15E+01 & 8.45E+00 & 6.63E+00 & 8.52E+00 & NaN   & 0.00E+00 \\
    \hline
    \multicolumn{2}{c|}{\textbf{Friedman Aligned Test }} & 4.1964 & 5.8036 & 5.3750 & 5.9464 & 3.2321 & 4.6964 & 4.0536 & \textbf{2.6964} \\
    \bottomrule
    \end{tabular}}
\end{table*}

\begin{table*}[htbp]
  \centering
  \caption{The mean value and the standard deviation of the objective during the 25 runs on the test instances C01 - C28 with $D$ = 30.}
  \label{tab:30d}
  \scalebox{0.7}[0.7]{
    \begin{tabular}{cc|c|c|c|c|c|c|c|c}
    \toprule
    \multicolumn{2}{c|}{Test Instances} & \multicolumn{1}{c|}{AGA-PPS} & \multicolumn{1}{c|}{LSHADE44+IDE} & \multicolumn{1}{c|}{LSHADE44} & \multicolumn{1}{c|}{UDE} & \multicolumn{1}{c|}{IUDE} & \multicolumn{1}{c|}{$\epsilon$MAg-ES} & \multicolumn{1}{c|}{C$^2$oDE} & \multicolumn{1}{c}{PPS-DE} \\
    \hline
        \multirow{2}[0]{*}{C01} & mean  & 7.10E-29 & 3.37E-11 & 1.02E-21 & 2.21E-15 & 4.13E-29 & 3.75E-28 & 6.34E-17 & \textbf{3.98E-29} \\
          & std   & 4.23E-29 & 4.11E-11 & 4.87E-21 & 7.08E-15 & 2.26E-29 & 7.20E-29 & 5.25E-17 & 2.42E-29 \\
    \multirow{2}[0]{*}{C02} & mean  & 6.27E-29 & 1.77E-11 & 2.86E-21 & 1.17E-14 & 4.42E-29 & 3.76E-28 & 6.88E-17 & \textbf{3.79E-29} \\
          & std   & 4.24E-29 & 2.52E-11 & 9.27E-21 & 3.65E-14 & 2.58E-29 & 7.26E-29 & 5.96E-17 & 2.32E-29 \\
    \multirow{2}[0]{*}{C03} & mean  & 1.08E+03 & 1.13E+07 & 1.12E+06 & 8.59E+01 & 1.29E+02 & \textbf{6.73E-28} & NaN   & 7.98E+01 \\
          & std   & 4.16E+02 & 4.60E+06 & 1.95E+06 & 2.29E+01 & 2.95E+01 & 1.07E-28 & NaN   & 1.73E+01 \\
    \multirow{2}[0]{*}{C04} & mean  & 2.19E+01 & 1.39E+01 & 1.97E+01 & 8.45E+01 & 1.36E+01 & 7.03E+01 & 1.10E+02 & \textbf{1.09E+00} \\
          & std   & 3.87E+00 & 7.78E-01 & 5.40E-01 & 2.36E+01 & 1.45E-06 & 3.11E+01 & 8.21E+00 & 3.76E+00 \\
    \multirow{2}[0]{*}{C05} & mean  & 6.47E-28 & 1.30E-16 & 4.25E-03 & 7.22E+00 & 5.71E-29 & \textbf{0.00E+00} & 4.27E-07 & 5.01E-29 \\
          & std   & 9.26E-28 & 7.82E-17 & 4.40E-03 & 1.07E+00 & 9.91E-29 & 0.00E+00 & 4.51E-07 & 5.88E-29 \\
    \multirow{2}[0]{*}{C06} & mean  & 4.09E+02 & 5.67E+03 & 3.96E+03 & 3.28E+02 & 4.29E+02 & 1.80E+02 & NaN   & \textbf{0.00E+00} \\
          & std   & 5.59E+01 & 1.03E+03 & 7.22E+02 & 1.05E+02 & 9.01E+01 & 9.96E+01 & NaN   & 0.00E+00 \\
    \multirow{2}[0]{*}{C07} & mean  & -2.21E+02 & -1.02E+01 & -5.55E+01 & -4.11E+02 & -3.27E+02 & \textbf{-7.01E+02} & NaN   & -2.45E+02 \\
          & std   & 6.65E+01 & 9.68E+01 & 1.08E+02 & 2.26E+02 & 1.14E+02 & 2.32E+02 & NaN   & 1.43E+02 \\
    \multirow{2}[0]{*}{C08} & mean  & \textbf{-2.84E-04} & -2.40E-04 & -2.80E-04 & -2.40E-04 & -2.80E-04 & -2.84E-04 & -2.06E-04 & 2.51E-02 \\
          & std   & 3.56E-09 & 4.05E-05 & 5.77E-10 & 4.94E-05 & 1.25E-12 & 3.94E-16 & 1.54E-05 & 9.57E-02 \\
    \multirow{2}[0]{*}{C09} & mean  & -2.67E-03 & \textbf{-2.67E-03} & \textbf{-2.67E-03} & \textbf{-2.67E-03} & \textbf{-2.67E-03} & -2.67E-03 & -2.66E-03 & \textbf{-2.67E-03} \\
          & std   & 8.85E-19 & 5.44E-09 & 1.33E-18 & 3.32E-16 & 0.00E+00 & 0.00E+00 & 7.83E-07 & 0.00E+00 \\
    \multirow{2}[0]{*}{C10} & mean  & -1.03E-04 & -9.00E-05 & -1.00E-04 & -9.12E-05 & -1.00E-04 & \textbf{-1.03E-04} & -7.18E-05 & 3.79E-02 \\
          & std   & 4.25E-09 & 8.64E-06 & 4.76E-10 & 1.79E-05 & 5.06E-15 & 0.00E+00 & 5.98E-06 & 1.87E-01 \\
    \multirow{2}[0]{*}{C11} & mean  & \textbf{-3.04E+02} & -8.55E-01 & -8.75E-01 & -2.70E+01 & -7.75E+00 & -9.25E-01 & NaN   & -6.43E-02 \\
          & std   & 3.06E+02 & 9.70E-02 & 1.10E-01 & 4.76E+00 & 7.36E+00 & 7.03E-15 & NaN   & 4.67E+00 \\
    \multirow{2}[0]{*}{C12} & mean  & 3.98E+00 & 6.07E+00 & 4.00E+00 & 1.57E+01 & \textbf{3.98E+00} & 4.61E+01 & 4.67E+00 & 4.98E+00 \\
          & std   & 4.26E-04 & 2.84E+00 & 1.35E-02 & 8.83E+00 & 1.07E-04 & 2.97E+01 & 2.06E-01 & 1.53E+00 \\
    \multirow{2}[0]{*}{C13} & mean  & 1.29E+01 & 3.27E+01 & 5.03E+01 & 9.64E+01 & 3.54E+00 & \textbf{2.89E-27} & 1.59E+01 & 5.43E-27 \\
          & std   & 3.02E+01 & 3.92E+01 & 1.36E+01 & 1.29E+02 & 1.61E+01 & 1.89E-27 & 3.28E+01 & 6.01E-27 \\
    \multirow{2}[0]{*}{C14} & mean  & 1.45E+00 & 1.93E+00 & 1.86E+00 & 1.59E+00 & \textbf{1.41E+00} & 1.63E+00 & 1.51E+00 & \textbf{1.41E+00} \\
          & std   & 6.09E-02 & 4.66E-02 & 4.47E-02 & 1.93E-01 & 1.05E-15 & 9.17E-02 & 3.93E-02 & 9.06E-16 \\
    \multirow{2}[0]{*}{C15} & mean  & \textbf{2.73E+00} & 1.29E+01 & 1.92E+01 & 9.27E+00 & 5.87E+00 & 6.75E+00 & 1.58E+01 & 6.38E+00 \\
          & std   & 1.38E+00 & 1.54E+00 & 3.61E+00 & 2.22E+00 & 3.55E+00 & 5.94E+00 & 3.57E+00 & 4.40E+00 \\
    \multirow{2}[0]{*}{C16} & mean  & \textbf{0.00E+00} & 1.56E+02 & 1.54E+02 & 8.92E+00 & 1.57E+00 & \textbf{0.00E+00} & 1.06E+01 & \textbf{0.00E+00} \\
          & std   & 0.00E+00 & 1.36E+01 & 1.53E+01 & 3.07E+00 & 5.12E-07 & 0.00E+00 & 3.25E+00 & 0.00E+00 \\
    \multirow{2}[0]{*}{C17} & mean  & 1.21E+00 & 1.03E+00 & 1.00E+00 & 1.03E+00 & \textbf{1.83E-01} & 9.72E-01 & NaN   & 4.57E-01 \\
          & std   & 3.17E-01 & 5.84E-03 & 1.82E-02 & 2.78E-03 & 2.78E-01 & 1.75E-02 & NaN   & 4.23E-01 \\
    \multirow{2}[0]{*}{C18} & mean  & 3.66E+01 & 7.54E+03 & 9.13E+03 & 9.84E+03 & 1.81E+02 & \textbf{3.65E+01} & NaN   & 7.07E+01 \\
          & std   & 1.39E-01 & 5.26E+03 & 6.63E+03 & 3.78E+03 & 4.83E+01 & 1.39E-01 & NaN   & 5.73E+01 \\
    \multirow{2}[0]{*}{C19} & mean  & \textbf{0.00E+00} & 1.28E-03 & 1.08E-03 & 1.97E+00 & \textbf{0.00E+00} & 7.60E+00 & NaN   & \textbf{0.00E+00} \\
          & std   & 0.00E+00 & 3.99E-04 & 9.40E-04 & 3.52E+00 & 0.00E+00 & 9.08E+00 & NaN   & 0.00E+00 \\
    \multirow{2}[0]{*}{C20} & mean  & 4.38E+00 & \textbf{2.92E+00} & 3.55E+00 & 4.00E+00 & 3.89E+00 & 7.66E+00 & 2.98E+00 & 3.65E+00 \\
          & std   & 6.63E-01 & 3.15E-01 & 2.21E-01 & 1.06E+00 & 2.83E-01 & 1.24E+00 & 5.51E-01 & 2.82E-01 \\
    \multirow{2}[0]{*}{C21} & mean  & \textbf{9.37E+00} & 2.77E+01 & 2.28E+01 & 1.25E+01 & 1.56E+01 & 4.84E+01 & 1.17E+01 & 1.99E+01 \\
          & std   & 6.49E+00 & 9.19E+00 & 8.99E+00 & 8.47E+00 & 1.09E+01 & 1.58E+01 & 4.20E+00 & 9.73E+00 \\
    \multirow{2}[0]{*}{C22} & mean  & 1.84E+02 & 1.18E+03 & 3.24E+03 & 2.21E+02 & 1.96E+01 & \textbf{2.47E-25} & NaN   & 1.59E-01 \\
          & std   & 2.09E+02 & 2.02E+03 & 3.17E+03 & 1.82E+02 & 3.50E+01 & 2.97E-26 & NaN   & 7.97E-01 \\
    \multirow{2}[0]{*}{C23} & mean  & 1.43E+00 & 1.91E+00 & 1.86E+00 & 1.50E+00 & 1.43E+00 & 1.65E+00 & 1.78E+00 & \textbf{1.42E+00} \\
          & std   & 4.48E-02 & 5.50E-02 & 6.09E-02 & 1.17E-01 & 3.79E-02 & 8.73E-02 & 2.20E-01 & 3.25E-02 \\
    \multirow{2}[0]{*}{C24} & mean  & 3.36E+00 & 1.42E+01 & 1.22E+01 & 9.27E+00 & \textbf{2.48E+00} & 9.14E+00 & 1.27E+01 & 3.24E+00 \\
          & std   & 1.50E+00 & 1.37E+00 & 1.04E+00 & 1.28E+00 & 6.28E-01 & 3.92E+00 & 3.33E+00 & 2.65E+00 \\
    \multirow{2}[0]{*}{C25} & mean  & 1.83E+01 & 1.48E+02 & 1.47E+02 & 1.59E+01 & 8.73E+00 & \textbf{0.00E+00} & 3.04E+01 & 4.40E+00 \\
          & std   & 7.25E+00 & 1.39E+01 & 1.27E+01 & 3.64E+00 & 4.95E+00 & 0.00E+00 & 1.09E+01 & 3.42E+00 \\
    \multirow{2}[0]{*}{C26} & mean  & 9.05E-01 & 1.03E+00 & 1.00E+00 & 1.03E+00 & \textbf{6.83E-01} & 9.78E-01 & NaN   & 7.94E-01 \\
          & std   & 1.92E-01 & 1.80E-03 & 2.11E-02 & 5.13E-03 & 2.45E-01 & 1.77E-02 & NaN   & 2.84E-01 \\
    \multirow{2}[0]{*}{C27} & mean  & 3.71E+01 & 4.16E+04 & 3.19E+04 & 3.07E+04 & 2.79E+02 & \textbf{3.66E+01} & NaN   & 1.90E+02 \\
          & std   & 1.83E+00 & 2.00E+04 & 1.13E+04 & 1.34E+04 & 5.92E+01 & 1.93E-01 & NaN   & 6.29E+01 \\
    \multirow{2}[0]{*}{C28} & mean  & 4.94E+01 & 1.55E+02 & 1.51E+02 & 6.50E+01 & 7.62E+01 & 5.84E+01 & NaN   & \textbf{6.34E+00} \\
          & std   & 2.17E+01 & 1.91E+01 & 2.04E+01 & 1.93E+01 & 2.95E+01 & 2.33E+01 & NaN   & 5.66E+00 \\
    \hline
    \multicolumn{2}{c|}{\textbf{Friedman Aligned Test }} & 3.3036 & 6.1071 & 5.6607 & 5.0714 & 3.1429 & 3.6071 &6.0357 &  \textbf{3.0714} \\
    \bottomrule
    \end{tabular}}
\end{table*}

\begin{table*}[htbp]
\centering
  \caption{The mean value and the standard deviation of the objective during the 25 runs on the test instances C01 - C28 with $D$ = 50.}
  \label{tab:50d}
  \scalebox{0.7}[0.7]{
    \begin{tabular}{cc|c|c|c|c|c|c|c|c}
    \toprule
    \multicolumn{2}{c|}{Test Instances} & \multicolumn{1}{c|}{AGA-PPS} & \multicolumn{1}{c|}{LSHADE44+IDE} & \multicolumn{1}{c|}{LSHADE44} & \multicolumn{1}{c|}{UDE} & \multicolumn{1}{c|}{IUDE} & \multicolumn{1}{c|}{$\epsilon$MAg-ES} & \multicolumn{1}{c|}{C$^2$oDE} & \multicolumn{1}{c}{PPS-DE} \\
    \hline
    \multirow{2}[0]{*}{C01} & mean  & 6.76E-25 & 1.21E-03 & 9.80E-19 & 6.77E-04 & \textbf{7.68E-28} & 2.87E-27 & 1.79E-05 & 8.73E-28 \\
          & std   & 8.43E-25 & 7.58E-04 & 1.88E-18 & 9.77E-04 & 6.72E-28 & 3.96E-28 & 1.56E-05 & 8.41E-28 \\
    \multirow{2}[0]{*}{C02} & mean  & 1.01E-24 & 8.25E-04 & 2.70E-17 & 2.89E-04 & \textbf{7.92E-28} & 2.89E-27 & 2.16E-05 & 8.79E-28 \\
          & std   & 3.71E-24 & 7.00E-04 & 7.75E-17 & 3.30E-04 & 6.41E-28 & 3.63E-28 & 1.74E-05 & 6.76E-28 \\
    \multirow{2}[0]{*}{C03} & mean  & 5.44E+03 & 4.14E+07 & 3.54E+06 & 3.41E+02 & 5.65E+02 & \textbf{4.06E-27} & NaN   & 1.04E+02 \\
          & std   & 1.40E+03 & 1.36E+07 & 5.08E+06 & 1.15E+02 & 1.93E+02 & 3.58E-28 & NaN   & 4.20E+01 \\
    \multirow{2}[0]{*}{C04} & mean  & 1.40E+02 & \textbf{1.40E+01} & 1.48E+02 & 1.61E+02 & 6.45E+01 & 1.19E+02 & 3.42E+02 & 3.00E+01 \\
          & std   & 2.67E+01 & 9.87E-01 & 7.43E+00 & 2.80E+01 & 2.20E+01 & 2.80E+01 & 1.36E+01 & 3.47E+01 \\
    \multirow{2}[0]{*}{C05} & mean  & 1.29E-19 & 4.31E-09 & 2.11E+01 & 3.19E+01 & 2.85E-28 & \textbf{0.00E+00} & 2.30E+01 & 3.96E-28 \\
          & std   & 3.98E-19 & 1.05E-08 & 3.80E-01 & 3.21E+00 & 1.77E-28 & 0.00E+00 & 8.91E-01 & 2.94E-28 \\
    \multirow{2}[0]{*}{C06} & mean  & 8.16E+02 & 8.99E+03 & 7.41E+03 & 6.56E+02 & 8.59E+02 & 2.87E+02 & NaN   & \textbf{1.59E+01} \\
          & std   & 8.44E+01 & 1.06E+03 & 1.20E+03 & 2.25E+02 & 1.25E+02 & 1.31E+02 & NaN   & 7.00E+01 \\
    \multirow{2}[0]{*}{C07} & mean  & -1.89E+02 & -3.65E+01 & -3.94E+01 & -6.73E+02 & -1.96E+02 & -1.37E+03 & \textbf{-6.30E+77} & -1.16E+02 \\
          & std   & 9.48E+01 & 1.21E+02 & 1.61E+02 & 2.44E+02 & 2.15E+02 & 3.40E+02 & 3.15E+78 & 1.98E+02 \\
    \multirow{2}[0]{*}{C08} & mean  & -1.17E-04 & 2.96E-04 & -1.30E-04 & 1.62E-03 & -1.30E-04 & \textbf{-1.35E-04} & NaN   & -1.32E-04 \\
          & std   & 3.21E-05 & 7.59E-05 & 2.33E-07 & 7.90E-04 & 4.28E-06 & 1.23E-16 & NaN   & 5.99E-06 \\
    \multirow{2}[0]{*}{C09} & mean  & -2.04E-03 & -1.56E-03 & \textbf{-2.04E-03} & \textbf{-2.04E-03} & \textbf{-2.04E-03} & 6.66E-01 & -1.45E-03 & 1.34E-02 \\
          & std   & 1.94E-09 & 2.35E-04 & 1.33E-18 & 5.84E-11 & 0.00E+00 & 1.87E+00 & 1.13E-04 & 3.41E-02 \\
    \multirow{2}[0]{*}{C10} & mean  & -4.75E-05 & 9.36E-05 & -4.82E-05 & 6.06E-05 & \textbf{-4.83E-05} & -4.83E-05 & 6.31E-04 & \textbf{-4.83E-05} \\
          & std   & 1.33E-06 & 3.77E-05 & 8.10E-08 & 4.70E-05 & 1.73E-11 & 1.83E-09 & 9.70E-05 & 1.95E-08 \\
    \multirow{2}[0]{*}{C11} & mean  & \textbf{-2.59E+03} & -7.30E-01 & -1.19E+00 & -9.48E+01 & -1.15E+03 & -3.70E+00 & 6.31E-04 & -4.84E+02 \\
          & std   & 3.64E+02 & 3.30E+00 & 2.44E+00 & 4.66E+01 & 1.16E+03 & 8.29E+00 & 9.70E-05 & 9.62E+02 \\
    \multirow{2}[0]{*}{C12} & mean  & 6.63E+00 & 7.36E+00 & 5.20E+01 & 1.25E+01 & 5.96E+00 & 5.06E+01 & 6.78E+00 & \textbf{5.44E+00} \\
          & std   & 4.07E+00 & 2.86E+00 & 2.09E+01 & 5.86E+00 & 1.51E+00 & 2.05E+01 & 5.25E-01 & 1.98E+00 \\
    \multirow{2}[0]{*}{C13} & mean  & 6.34E+01 & 9.14E+01 & 6.50E+02 & 1.37E+03 & 1.98E+01 & 2.95E+02 & NaN   & \textbf{3.46E-26} \\
          & std   & 5.42E+01 & 2.49E+01 & 1.02E+02 & 4.17E+02 & 4.05E+01 & 4.44E+02 & NaN   & 2.37E-26 \\
    \multirow{2}[0]{*}{C14} & mean  & 1.17E+00 & 1.49E+00 & 1.41E+00 & 1.29E+00 & 1.10E+00 & 1.34E+00 & 1.46E+00 & \textbf{1.10E+00} \\
          & std   & 8.75E-02 & 2.97E-02 & 2.96E-02 & 9.74E-02 & 6.80E-16 & 3.74E-02 & 6.62E-02 & 6.80E-16 \\
    \multirow{2}[0]{*}{C15} & mean  & \textbf{5.25E+00} & 1.45E+01 & 1.78E+01 & 1.17E+01 & 6.00E+00 & 1.45E+01 & NaN   & 6.63E+00 \\
          & std   & 1.26E+00 & 1.65E+00 & 3.00E+00 & 1.43E+00 & 5.10E+00 & 1.01E+01 & NaN   & 4.96E+00 \\
    \multirow{2}[0]{*}{C16} & mean  & 6.28E-02 & 2.72E+02 & 2.72E+02 & 1.26E+01 & 6.28E+00 & \textbf{0.00E+00} & 5.47E+01 & \textbf{0.00E+00} \\
          & std   & 3.14E-01 & 1.77E+01 & 1.84E+01 & 7.25E-15 & 2.70E-05 & 0.00E+00 & 1.50E+01 & 0.00E+00 \\
    \multirow{2}[0]{*}{C17} & mean  & 1.01E+00 & 1.05E+00 & 1.04E+00 & 1.05E+00 & \textbf{6.09E-01} & 1.03E+00 & NaN   & 1.30E+00 \\
          & std   & 2.75E-01 & 5.86E-04 & 5.57E-03 & 1.56E-03 & 2.27E-01 & 6.12E-03 & NaN   & 4.25E-01 \\
    \multirow{2}[0]{*}{C18} & mean  & \textbf{3.66E+01} & 2.00E+04 & 2.05E+04 & 3.40E+04 & 3.73E+02 & 3.66E+01 & NaN   & 2.27E+02 \\
          & std   & 3.74E-01 & 6.83E+03 & 7.21E+03 & 9.62E+03 & 3.77E+01 & 5.93E-01 & NaN   & 9.00E+01 \\
    \multirow{2}[0]{*}{C19} & mean  & \textbf{0.00E+00} & 3.54E-02 & 6.66E-02 & 6.42E+00 & 7.06E-01 & 1.25E+01 & NaN   & \textbf{0.00E+00} \\
          & std   & 0.00E+00 & 1.93E-02 & 3.89E-02 & 7.26E+00 & 2.45E+00 & 9.73E+00 & NaN   & 0.00E+00 \\
    \multirow{2}[0]{*}{C20} & mean  & 1.03E+01 & \textbf{5.63E+00} & 8.12E+00 & 7.85E+00 & 8.68E+00 & 1.52E+01 & 1.30E+01 & 8.47E+00 \\
          & std   & 6.01E-01 & 2.93E-01 & 2.99E-01 & 1.64E+00 & 3.99E-01 & 5.51E-01 & 3.67E-01 & 3.16E-01 \\
    \multirow{2}[0]{*}{C21} & mean  & \textbf{6.62E+00} & 6.28E+01 & 6.53E+01 & 7.64E+00 & 8.73E+00 & 5.53E+01 & 4.45E+01 & 1.30E+01 \\
          & std   & 3.77E+00 & 1.43E+00 & 2.04E+00 & 4.22E+00 & 5.30E+00 & 1.68E+01 & 1.29E+01 & 7.58E+00 \\
    \multirow{2}[0]{*}{C22} & mean  & 4.12E+03 & 1.13E+04 & 1.45E+04 & 4.09E+03 & 5.39E+02 & 9.76E+02 & NaN   & \textbf{1.89E+01} \\
          & std   & 6.42E+03 & 6.03E+03 & 7.73E+03 & 3.05E+03 & 5.01E+02 & 6.06E+02 & NaN   & 2.73E+01 \\
    \multirow{2}[0]{*}{C23} & mean  & 1.15E+00 & 1.44E+00 & 1.42E+00 & 1.26E+00 & 1.11E+00 & 1.34E+00 & 1.57E+00 & \textbf{1.11E+00} \\
          & std   & 3.99E-02 & 2.98E-02 & 3.13E-02 & 7.70E-02 & 1.74E-02 & 4.52E-02 & 2.56E-02 & 1.74E-02 \\
    \multirow{2}[0]{*}{C24} & mean  & 5.50E+00 & 1.56E+01 & 1.43E+01 & 1.14E+01 & 4.24E+00 & \textbf{1.88E+00} & 1.82E+01 & 2.48E+00 \\
          & std   & 9.07E-01 & 1.57E+00 & 1.28E+00 & 1.38E+00 & 3.01E+00 & 1.49E+01 & 1.92E+00 & 6.28E-01 \\
    \multirow{2}[0]{*}{C25} & mean  & 5.30E+01 & 2.65E+02 & 2.53E+02 & 2.34E+01 & 6.85E+00 & \textbf{0.00E+00} & 1.17E+02 & 1.60E+01 \\
          & std   & 1.67E+01 & 2.00E+01 & 1.69E+01 & 7.59E+00 & 1.75E+00 & 0.00E+00 & 3.40E+01 & 5.89E+00 \\
    \multirow{2}[0]{*}{C26} & mean  & 9.91E-01 & 1.05E+00 & 1.04E+00 & 1.05E+00 & \textbf{9.82E-01} & 1.03E+00 & NaN   & 9.98E-01 \\
          & std   & 1.50E-01 & 3.46E-03 & 3.29E-03 & 3.76E-03 & 5.46E-02 & 7.26E-03 & NaN   & 1.28E-01 \\
    \multirow{2}[0]{*}{C27} & mean  & 4.07E+01 & 7.60E+04 & 8.40E+04 & 1.09E+05 & 4.95E+02 & \textbf{3.65E+01} & NaN   & 4.02E+02 \\
          & std   & 1.93E+01 & 2.03E+04 & 2.88E+04 & 1.88E+04 & 5.22E+01 & 5.16E-06 & NaN   & 7.99E+01 \\
    \multirow{2}[0]{*}{C28} & mean  & 1.39E+02 & 2.74E+02 & 2.67E+02 & 1.33E+02 & 1.84E+02 & 9.53E+01 & NaN   & \textbf{2.50E+01} \\
          & std   & 3.65E+01 & 1.88E+01 & 1.78E+01 & 2.19E+01 & 2.69E+01 & 4.43E+01 & NaN   & 9.59E+00 \\
    \hline
    \multicolumn{2}{c|}{\textbf{Friedman Aligned Test}} & 3.3214 & 6.1429 & 5.7321 & 5.125 & 2.9464 & 3.5714 & 6.4643 & \textbf{2.6964} \\
    \bottomrule
    \end{tabular}}
\end{table*}

\section{Conclusion}
\label{sec:con}
This paper extended the PPS framework to solve CSOPs. More specifically, the proposed PPS-DE integrated PPS technique and an adaptive DE algorithm to deal with CSOPs. Three trial vector generation strategies --- DE /rand/1, DE/current-to-rand/1, and DE/current-to-pbest/1 are used in the proposed PPS-DE. In PPS-DE, two sub-populations are employed to collaborated with each other to search for global optimal solutions. The top sub-population adopts the PPS technique to deal with constraints, while the bottom sub-population use the SF technique to deal with constraints. In the top sub-population, all the three trial vector generation strategies are used to generate offsprings.  In the bottom sub-population, a strategy adaptation, in which the trial vector generation strategies are periodically self-adapted by learning from their experiences in generating promising solutions in the top sub-population, is employed to choose a suitable trial vector generation strategy in each generation. Furthermore, the parameter adaptation principle of LSHADE44 is employed in both sup-populations in the proposed PPS-DE. In the push stage of PPS-DE, a CSOP is optimized without considering any constraints, which can help PPS-DE to cross infeasible regions in front of the global optimum. In the pull stage of PPS-DE, the CSOP is optimized with an improved epsilon constraint-handling method. The comprehensive experiments indicate that the proposed PPS-DE achieves significantly better results than the other seven constrained DEs on most of the benchmark problems provided in the CEC2018 competition on real parameter single objective optimization.

It is also worthwhile to point out that PPS technique is not only a powerful constraint-handling method but also a general search framework which is focus on solving optimization problems with constraints. Obviously, a lot of work need to be done to improve the performance of PPS-DE, such as, the enhanced constraint-handling mechanisms in the pull stage, the enhanced strategies to switch the search behavior, and the machine learning approaches integrated in the PPS framework. For another future work, the proposed PPS will be applied to solve constrained optimization problems with more than three objectives, i.e., constrained many-objective optimization problems, to further verify the effect of PPS. Some real-world optimization problems will also be used to test the performance of the PPS embedded in different DE variants.

\section*{Acknowledgement}
This research work was supported by the Key Lab of Digital Signal and Image Processing of Guangdong Province, the National Natural Science Foundation of China under Grant (61175073, 61300159, 61332002, 51375287), the Natural Science Foundation of Jiangsu Province of China under grant SBK2018022017, China Postdoctoral Science Foundation under grant 2015M571751, and Project of International, as well as Hongkong, Macao\&Taiwan Science and Technology Cooperation Innovation Platform in Universities in Guangdong Province (2015KGJH2014).



\section*{Reference}
\bibliographystyle{elsarticle-num}
\bibliography{pps_de.bib}





\end{document}